\documentclass[a4paper,USenglish,cleveref, autoref]{lipics-v2019}
\usepackage{bm}
\usepackage[utf8]{inputenc}
\usepackage{url}
\usepackage{amssymb}
\usepackage{amsfonts}
\usepackage{graphicx}
\usepackage{url}
\usepackage{lstsemantic}
\usepackage{lstlangmizar}
\usepackage{booktabs}
\usepackage{footnoterange}

\usepackage{float}
\floatstyle{boxed}

\usepackage{hyperref}

\title{Hammering Mizar by Learning Clause Guidance}

\author{Jan Jakub\r{u}v}{Czech Technical University in Prague}{}{[orcid]}{Supported by the \textit{AI4REASON} ERC Consolidator grant number 649043.}

\author{Josef Urban}{Czech Technical University in Prague}{}{[orcid]}{Supported by the \textit{AI4REASON} ERC Consolidator grant number 649043, and by the Czech project AI\&Reasoning CZ.02.1.01/0.0/0.0/15\_003/0000466 and the European Regional Development Fund.}

\authorrunning{Jakub\r{u}v, Urban}

\titlerunning{Hammering Mizar by Learning Clause Guidance}

\Copyright{Jakub\r{u}v and Urban}

\ccsdesc{Artificial intelligence, Machine learning, Knowledge representation and reasoning}

\keywords{Proof automation, ITP hammers, Automated theorem proving, Machine learning}

\nolinenumbers

\hideLIPIcs

\newcommand*{\M}{\mathcal{M}}

\renewcommand*{\S}{\mathcal{S}}
\renewcommand*{\P}{\mathcal{P}}
\newcommand*{\T}{\mathcal{T}}

\begin{document}

\maketitle{}

\begin{abstract}
  We describe a very large improvement of existing hammer-style proof automation over large ITP libraries by combining learning and theorem proving.  In particular, we have integrated state-of-the-art machine learners into the E automated theorem prover, and developed methods that allow learning and efficient internal guidance of E over the whole Mizar library. The resulting trained system improves the real-time performance of E on the Mizar library by  70\% in a single-strategy setting.

\end{abstract}

\section{Introduction}
\label{Intro}
Proof automation for
interactive theorem provers (ITPs) has been a major factor behind the recent progress in formal verification.
In particular, \emph{Hammers} linking  ITPs
with automated theorem provers
(ATPs) produce a major speed\-up of formalization~\cite{hammers4qed}.
The main AI component of existing hammers has so far been
\emph{premise selection}~\cite{abs-1108-3446}, where only the most
relevant facts are chosen from the large ITP libraries as axioms for
proving a new conjecture. Machine learning from the large number of
proofs in the ITP libraries has resulted in the strongest premise
selection methods
~\cite{abs-1108-3446,holyhammer,KaliszykU13b,FarberK15,BlanchetteGKKU16,IrvingSAECU16-short}.
Premise selection however does not guide the theorem proving processes once the premises are selected.
The success of machine learning in the high-level premise selection task has motivated development of low-level \emph{internal
  proof search guidance}. This has been recently started both for
ATPs~\cite{UrbanVS11,KaliszykU15,JakubuvU17a,LoosISK17,FarberKU17} and
also in the context of tactical ITPs~\cite{GauthierKU17,GransdenWR15}.

Recently, we have added~\cite{abs-1903-03182} two state-of-the-art machine learning methods
to the ENIGMA~\cite{JakubuvU17a,JakubuvU18} algorithm that efficiently
guides saturation-style proof search in ATPs such as E
\cite{Sch02-AICOMM,Schulz13}. The first method trains gradient boosted trees on
efficiently extracted manually designed (handcrafted) clause
features. The second method uses end-to-end training of recursive
neural networks, thus removing the need for handcrafted features
. While the second method seems very promising and already improves on
a simpler linear classifier when used for guidance, its efficient
training and use over a large ITP library is still practically
challenging. On the other hand, our recent experiments with efficient
\emph{feature hashing} have shown that the very good performance of
gradient boosted trees is maintained even after significant
dimensionality reduction of the feature set~\cite{abs-1903-03182}. This opens the way to
training learning-based internal guidance of saturation search even on
very large ITP libraries, where the hundreds of thousands of handcrafted features would otherwise make the trained guiding systems impractically slow.

In this work we conduct the first practical evaluation of learning-based
internal guidance of state-of-the-art saturation provers such as E in
a realistic large-library hammer setting, with realistic time limits.
The results turn out to be unexpectedly good, 
improving the real-time performance of E on the whole Mizar Mathematical Library (MML)~\cite{mizar-in-a-nutshell} by 70\% in a
single-strategy setting. We believe that this is a breakthrough that will quickly lead to ubiquitous deployment
of 
ATPs equipped with learning-based internal guidance in large-theory theorem proving and in hammer-style
ITP assistance.

The rest of the paper is organized as
follows. Section~\ref{sec:guidance} summarizes the general
saturation-style ATP setting and explains how machine learning can be
trained and used over a large library of problems to guide the
saturation search. Section~\ref{sec:hand} discusses the practical
implementation of ENIGMA, i.e., the features, classifiers, and the
feature hashing used to make the ENIGMA guidance both strong and
efficient on a large library. Section~\ref{Experiments} is our main
contribution. We evaluate the latest ENIGMA on the whole Mizar Mathematical
Library and show that in several iterations of proving and learning we can develop very strong strategies and solve
in low time limits many previously unsolved problems.

\section{Enhancing ATPs with Machine Learning}
\label{sec:guidance}

\textbf{Automated Theorem Proving.} 
State-of-the-art saturation-based automated theorem provers (ATPs) for
first-order logic (FOL), such as E \cite{Sch02-AICOMM} and Vampire~\cite{Vampire}
are today's most advanced tools for general reasoning across a variety of
mathematical and scientific domains. 
Many ATPs employ the \emph{given clause algorithm}, translating
the input FOL problem $T\cup\{\lnot C\}$ into a refutationally
equivalent set of clauses.
The search for a contradiction is performed maintaining sets of
\emph{processed} ($P$) and \emph{unprocessed} ($U$) clauses.
The algorithm repeatedly selects a \emph{given clause} $g$ from $U$,
moves $g$ to $P$, and extends $U$ with all clauses inferred with $g$ and $P$.
This process continues until a contradiction is found, $U$ becomes empty, or 
a resource limit is reached.
The search space of this loop grows quickly and it is a well-known fact that the
selection of the right given clause is crucial for success.
Machine learning from a large number of proofs and proof searches may help guide
the selection of the given clauses.

E allows the user to select a \emph{proof search strategy} $\S$ to guide the
proof search.
An E strategy $\S$ specifies parameters such as term ordering, literal selection
function, clause splitting, paramodulation setting, premise selection, and, most
importantly for us, the \emph{given clause selection} mechanism.
The given clause selection in E is implemented using a collection of \emph{weight
functions}.
These weight functions are used in a round robin manner to select the given
clause.

\textbf{Machine Learning of Given Clause Selection.}
To facilitate machine learning research, E implements an option
under which each successful proof search gets analyzed and 
the prover outputs a list of clauses annotated as either \emph{positive} or \emph{negative} training examples.
Each processed clause which is present in the final proof is classified as
positive. %
On the other hand, processing of clauses not present in the final proof was redundant,
hence they are classified as negative.
Our goal is to learn such classification (possibly conditioned on the
problem and its features) in a way that generalizes and allows solving
related problems.

Given a set of problems $\P$, we can run E with a strategy $\S$ and obtain positive and
negative training data
$\T$ from each of the successful proof searches.
Various machine learning methods can be used to learn the clause classification
given by $\T$, each method yielding a \emph{classifier} or \emph{model} $\M$.
In order to use the model $\M$ in E, $\M$ needs to provide the function to
compute the weight of an arbitrary clause.
This weight function is then used to guide future E runs.

\textbf{Guiding ATPs with Learned Models.}
A model $\M$ can be used in E in different ways.
We use two methods to combine $\M$ with a strategy $\S$.
Either (1) we use $\M$ to select \emph{all} the given clauses, or (2) we combine $\M$
with the given clause guidance from $\S$ so that roughly half of the clauses are
selected by $\M$.
Proof search settings other than given clause guidance are inherited from $\S$.
We denote the resulting E strategies as (1) $\S\odot\M$, and (2) $\S\oplus\M$.

\section{EN\underline{IGM}A: Inference Guiding Machine}
\label{sec:hand}

\textbf{Machine Learning in Practice.}
ENIGMA~\cite{JakubuvU17a,JakubuvU18} is our \emph{efficient}
learning-based method for guiding given clause selection in saturation-based
ATPs, implementing the framework suggested in the previous
Section~\ref{sec:guidance}.
First-order clauses need to be represented in a format recognized by the
selected learning method.
While neural networks have been very recently practically used for
internal guidance with ENIGMA~\cite{abs-1903-03182}, the strongest setting currently uses
manually engineered \emph{clause features} and fast non-neural state-of-the-art
gradient boosted trees library~\cite{DBLP:conf/kdd/ChenG16}.

\textbf{Clause Features.}
Clause features represent a finite set of various syntactic properties of
clauses, and are used to encode clauses by a fixed-length numeric vector.
Various machine learning methods can handle numeric vectors and their success
heavily depends on the selection of correct clause features.
Various possible choices of efficient clause features for theorem prover
guidance have been experimented
with~\cite{JakubuvU17a,JakubuvU18,KaliszykUMO18,DBLP:conf/ijcai/KaliszykUV15}.
The original ENIGMA~\cite{JakubuvU17a} uses term-tree walks of
length 3 as features, while the second version~\cite{JakubuvU18} reaches better
results by employing various additional features.

Since there are only finitely many features in any training data, the features
can be serially numbered.
This numbering is fixed for each experiment.
Let $n$ be the number of different features appearing in the training data.
A clause $C$ is translated to a feature vector
$\varphi_C$ whose $i$-th member counts the number of occurrences of the $i$-th 
feature in $C$.
Hence every clause is represented by a sparse numeric vector of length $n$.
Additionally, we embed information about the conjecture currently being proved
in the feature vector, yielding vectors of length $2n$.
See \cite{abs-1903-03182,JakubuvU18} for more details.

\textbf{From Logistic Regression to Decision Trees.}
So far, the development of ENIGMA was focusing on fast and practically usable
methods, allowing E users to directly benefit from our work. 
Simple but fast linear classifiers such as \emph{linear SVM} and \emph{logistic
regression} efficiently implemented by the LIBLINEAR open source
library~\cite{DBLP:journals/jmlr/FanCHWL08} were used in our initial
experiments~\cite{JakubuvU18}.
Our recent experiments~\cite{abs-1903-03182} report improved performance with
\emph{gradient boosted trees}, while maintaining efficiency.
Gradient boosted trees are ensembles of decision trees trained by tree boosting.
In particular, we use their implementation in the XGBoost
library~\cite{DBLP:conf/kdd/ChenG16}. 

The model $\M$ produced by XGBoost consists of a set (\emph{ensemble}~\cite{Polikar06}) of decision trees.
The inner nodes of the decision trees consist of conditions on feature values,
while the leafs contain numeric scores.
Given a vector $\varphi_C$ representing a clause $C$, each tree in $\M$ is navigated
to the unique leaf using the values from $\varphi_C$, and the corresponding leaf scores
are aggregated across all trees.
The final score is translated to yield the probability that $\varphi_C$ represents a
positive clause.
When using $\M$ as a weight function in E, the probabilities are turned into binary
classification, assigning weight $1.0$ for probabilities $\ge 0.5$ and
weight $10.0$ otherwise.
Our experiments with scaling of the weight by the probability did not yet yield
improved functionality.

\textbf{Feature Hashing.}
The vectors representing clauses have so far had length $n$ when $n$
is the total number of features in the training data $\T$ (or $2n$
with conjecture features).  Experiments revealed that XGBoost is capable of
dealing with vectors up to the length of $10^5$ with a reasonable performance.
This might be enough for
smaller benchmarks but with the need to train on bigger training data,
we might need to handle much larger feature sets.  In experiments with
the whole translated Mizar Mathematical Library, the feature vector
length can easily grow over $10^6$.  This significantly increases both
the training and the clause evaluation times.  To handle such larger
data sets, we have implemented a simple \emph{hashing} method to
decrease the dimension of the vectors.

Instead of serially numbering all features, %
we represent
each feature $f$ by a unique string and apply a general-purpose string
hashing function%
\footnote{We use the following hashing function $\mathit{sdbm}$:
$ h_i = s_i + (h_{i-1} \ll 6) + (h_{i-1} \ll 16) - h_{i-1} $.
}
to obtain a number $n_f$ within a required range (between 0 and an
adjustable \emph{hash base}).
The value of $f$ is then stored in the feature vector at the position $n_f$.
If different features get mapped to the same
vector index, the corresponding values are summed up.
See \cite{abs-1903-03182} for more details.

\section{Experiments}
\label{Experiments}

\begin{table}[t]
\caption{Number of Mizar problems solved in 10 seconds by various ENIGMA strategies.}
\label{tab:reallife}
\vspace{-6mm}
\setlength\tabcolsep{1mm}
\begin{center}
\begin{tabular}{l|c|cc|cc|cc|cc}
	    & $\S$ & $\S\odot\M_9^0$ & $\S\oplus\M_9^0$ & $\S\odot\M^1_9$ & $\S\oplus\M^1_9$ & $\S\odot\M^2_9$ & $\S\oplus\M^2_9$ & $\S\odot\M^3_9$ & $\S\oplus\M^3_9$ \\
\hline
solved &    14933  & 16574 & 20366 & 21564 & 22839 & 22413 & 23467 & 22910 & 23753 \\
$\S\%$ & +0\% & +10.5\% & +35.8\% & +43.8\% & +52.3\% & +49.4\% & +56.5\% &
+52.8\% & +58.4 \\
$\S+$ &  +0 & +4364 & +6215 & +7774 & +8414 & +8407 &  +8964 & +8822 & +9274 \\
$\S-$ & -0 &   -2723 & -782 & -1143 & -508 & -927 & -430   & -845 & -454 \\
\end{tabular}
\end{center}
\begin{center}
\begin{tabular}{l|cc|cc}
	    & $\S\odot\M^3_{12}$ & $\S\oplus\M^3_{12}$ & $\S\odot\M^3_{16}$ & $\S\oplus\M^3_{16}$ \\
\hline
solved &    24159  & 24701  & 25100 & 25397 \\
$\S\%$ & +61.1\% & +64.8\% & +68.0\% & +70.0\% \\
$\S+$ &  +9761 & +10063 & +10476 & +10647 \\
$\S-$ & -535   &-295  & -309 & -183  \\
\end{tabular}
\end{center}
\end{table}

\begin{table}[t]
\begin{small}
\caption{Comparison of several developed strategies in higher time limits.}
\label{tab:higher}
\vspace{-6mm}
\setlength\tabcolsep{1mm}
\begin{center}
\begin{tabular}{lcccccc}
	    & $\S$ (30s) & $\S\oplus\M^2_9$ (30s) &  $\S\oplus\M^2_9$ (60s) & $\S\oplus\M^3_9$ (60s) & $\S\oplus\M^3_{12}$ (60s) & $\S\oplus\M^3_{16}$ (60s) \\
\toprule
solved &    15554 & 24154 & 24495  & 24762 & 25540 & 26107 \\ 
hard &  75 & 891 & 956 & 1017 & 1192 & 1296 \\
\end{tabular}
\end{center}
\end{small}
\end{table}

\begin{table}[t]
\caption{Training statistics and inference speed for different tree depths.}
\label{tab:training}
\vspace{-6mm}
\center
\begin{tabular}{lccccc}
Tree depth & training error & real time & CPU time & model size (MB) & inference speed \\ \midrule
9  & 0.201 & 2h41m & 4d20h  & 5.0 & 5665.6 \\ 
12 & 0.161 & 4h12m & 8d10h  & 17.4 & 4676.9 \\
16 & 0.123 & 6h28m & 11d18h & 54.7 & 3936.4  \\ \bottomrule
\end{tabular}
\end{table}

\begin{table}[t]
\caption{Effect of looping on 10k randomly selected problems.}
\vspace{-6mm}
\label{tab:looping}
\setlength\tabcolsep{1mm}
\begin{center}
\begin{tabular}{l|c|c|c|c|c|c|c|c}
	    & $\S$ & $\S\oplus\M^0$ & $\S\oplus\M^1$ & $\S\oplus\M^2$ & $\S\oplus\M^3$ & $\S\oplus\M^4$ & $\S\oplus\M^5$ & $\S\oplus\M^6$ \\
\hline
solved & 2487 & 3204 & 3625 & 3755 & 3838 & 3854 & 3892 & 3944 
\\
$\S\%$ &+0\% & +28.8\% & +45.7\% & +50.9\% & +54.3\% & +54.9\% & +56.4\% & +58.5\%
\end{tabular}
\end{center}
\end{table}

The experiments are done on a large benchmark of $57880$
Mizar40~\cite{KaliszykU13b} problems\footnote{\url{http://grid01.ciirc.cvut.cz/~mptp/7.13.01_4.181.1147/MPTP2/problems_small_consist.tar.gz}} from the MPTP
dataset~\cite{Urban06}. Since we are here interested in internal guidance rather than in premise selection, we 
have used the small
  (\emph{bushy}, re-proving) versions of the problems, however without previous ATP
  minimization. 
We start with a good evolutionarily
optimized~\cite{JakubuvU17} E strategy $\S$ that performed best in previous experiments on the smaller MPTP2078 dataset.
We run $\S$ for 10s on the whole library, producing the first proofs, we learn from them the next guiding strategy, and this is iterated with the growing body of proofs.
All problems are run on the same hardware\footnote{Intel(R) Xeon(R)
  CPU E5-2698 v3 @ 2.30GHz with 256G RAM.} and with the same memory
limits employing multiple cores (around 300) for massive parallel evaluation.

Table~\ref{tab:reallife} shows the number of Mizar problems solved in 10 seconds by the baseline strategy $\S$ and by each
iteration of learning and proving with the learned guidance.
The model $\M_9^0$ is trained on the training data coming from the problems solved
by $\S$ with the maximum depth of XGBoost decision trees set to 9.
We further \emph{loop} this process and models $\M_9^n$ are trained on all the
problems solved by $\S$, and by all the previous $\S\odot\M_9^k$ and
$\S\oplus\M_9^k$ for $k<n$.
Models $\M_{12}^3$ and $\M_{16}^3$ are trained on the same data as $\M_9^3$ but
with the tree depth increased to 12 and 16.
XGBoost models contain 200 decision trees and the hash base is set to $2^{15}$.
In the row $\S\%$ we show the percentage gain over the baseline strategy $\S$, while 
$\S+$ and $\S-$ are the additions and missing solutions w.r.t. $\S$.
We can see that new problems are added with every iteration of looping.
Combined versions ($\oplus$) typically perform better and lose less solutions.
Increasing the tree depth to 16 leads to a strategy that outperforms the baseline by
rather astonishing 70\%.

Table~\ref{tab:higher} compares several of our new strategies with higher time limits and also shows the number of \emph{hard} problems, i.e., the problems unsolved by any method developed previously in~\cite{KaliszykU13b}. Our best strategy $S\oplus\M^3_{16}$ solves 26107 problems in 60s. Note that the 60s portfolio of our six best previous evolutionarily developed strategies for Mizar (i.e., each run for 10s) solves only 22068 problems, i.e., the single new strategy is 18.3\% better. Vampire in the CASC (best portfolio) mode run in 300s has solved 27842 of these problems in 300s in~\cite{KaliszykU13b}.

Table~\ref{tab:training} shows the training times, model
sizes and inference speeds of XGBoost in the 4th iteration of proving
and learning, using different tree depths.  The training data is a
sparse matrix with 65536 ($=2*2^{15}$) columns (features) consisting of 63M
examples. The total number of non-empty entries in the matrix is 5B
(40GB).  The inference speed is the average of the generated clauses
per second measured on problems that timed out in all three runs. Note
that despite the decrease of the inference speed with the more
complicated XGBoost models, their accuracy and real-time performance
grows (cf. Table~\ref{tab:higher}). Training of better models on the
millions of proof search examples however already requires significant
resources - almost 12 CPU days for the best model with tree depth 16.

Table~\ref{tab:looping} presents additional shorter experiments with more looping performed
on a randomly selected 10k problems. %
The tree depth is set to 9.
Again, the model $\M^0$ is trained only on the problems solved by $\S$ and the next
models are obtained by looping.
The highest improvement is achieved after the first learning ($\M^0$), 
however, the next iterations do not stop to add improvements.

\section{Conclusion and Future Work}
We have taken a good previously tuned E strategy and turned it into a
learning-guided strategy that is 70\% stronger in real time.  We have
done that by several iterations of MaLARea-style~\cite{US+08} feedback loop between
proving and learning over a large mathematical library. The iterations
here are however not done for learning premise selection as in
MaLARea, but for learning efficient internal guidance.  While
developing this kind of efficient internal guidance for
state-of-the-art saturation ATPs has been challenging and took time,
the very large gains obtained here show that this has been very well
invested effort. Future work will certainly focus on even stronger
learning methods and also on more dynamic proof state characterization
such as ProofWatch~\cite{DBLP:conf/itp/GoertzelJ0U18} and ENIGMAWatch~\cite{ProofWatch_Meets_ENIGMA_First}. It is however clear that this is the point when machine
learning guidance has very strongly overtaken the human
development of ATP strategies over large problem corpora.

\begin{small}
\bibliographystyle{abbrv}
\bibliography{ate11.bib,stsbib.bib,itp.bib}
\end{small}

\appendix

\section{The Starting Strategy $\S$ Used in the Experiments}
\label{sec:app}

The following E strategy has been used to undertake the experimental evaluation
in Section~\ref{Experiments}.  The given clause selection strategy
(heuristic) is defined using the parameter ``\verb+-H+''.

\begin{verbatim}
--definitional-cnf=24 --split-aggressive --simul-paramod -tKBO6 -c1 -F1
-Ginvfreq -winvfreqrank --forward-context-sr --destructive-er-aggressive 
--destructive-er --prefer-initial-clauses -WSelectMaxLComplexAvoidPosPred
-H'(1*ConjectureTermPrefixWeight(DeferSOS,1,3,0.1,5,0,0.1,1,4),
    1*ConjectureTermPrefixWeight(DeferSOS,1,3,0.5,100,0,0.2,0.2,4),
    1*Refinedweight(ConstPrio,4,300,4,4,0.7),
    1*RelevanceLevelWeight2(PreferProcessed,0,1,2,1,1,1,200,200,2.5, 
                                                           9999.9,9999.9),
    1*StaggeredWeight(DeferSOS,1),
    1*SymbolTypeweight(DeferSOS,18,7,-2,5,9999.9,2,1.5),
    2*Clauseweight(ConstPrio,20,9999,4),
    2*ConjectureSymbolWeight(DeferSOS,9999,20,50,-1,50,3,3,0.5),
    2*StaggeredWeight(DeferSOS,2))'
\end{verbatim}

\end{document}